\documentclass[lettersize,journal]{IEEEtran}

 \PassOptionsToPackage{numbers, compress, authoryear}{natbib}
\usepackage[colorlinks=true,linkcolor=blue,citecolor=blue]{hyperref}%
\usepackage{url}                    
\usepackage{booktabs}           
\usepackage{amsfonts}           
\usepackage{nicefrac}           
\usepackage{microtype}          
\usepackage{longtable}
\usepackage{graphicx}
\usepackage{subcaption}
\usepackage{algorithm}
\usepackage{capt-of}
\usepackage{xcolor}
\usepackage{mathtools}        
\usepackage{algpseudocode}
\newcommand{\dataset}[1]{\textsc{#1}}

\newlength\savewidth

\definecolor{myblue1}{RGB}{0,177,234}
\newcommand{\multiline}[3][11mm]{\begin{minipage}[c][#1]{#2}#3\end{minipage}}

\title{Multi-Grid Graph Neural Networks with Self-Attention for Computational Mechanics}

\author{
  Paul Garnier, Jonathan Viquerat, Elie Hachem\thanks{Correspondence: \texttt{elie.hachem@minesparis.psl.eu}} \\
  MINES Paristech - PSL Research University\\
  CEMEF \\
}

\begin{document}

\maketitle

\begin{abstract}
Advancement in finite element methods have become essential in various disciplines, and in particular for Computational Fluid Dynamics (CFD), driving research efforts for improved precision and efficiency. While Convolutional Neural Networks (CNNs) have found success in CFD by mapping meshes into images, recent attention has turned to leveraging Graph Neural Networks (GNNs) for direct mesh processing. This paper introduces a novel model merging Self-Attention with Message Passing in GNNs, achieving a 15\% reduction in RMSE on the well known flow past a cylinder benchmark. Furthermore, a dynamic mesh pruning technique based on Self-Attention is proposed, that leads to a robust GNN-based multigrid approach, also reducing RMSE by 15\%. Additionally, a new self-supervised training method based on BERT is presented, resulting in a 25\% RMSE reduction. The paper includes an ablation study and outperforms state-of-the-art models on several challenging datasets, promising advancements similar to those recently achieved in natural language and image processing. Finally, the paper introduces a dataset with meshes larger than existing ones by at least an order of magnitude. Code and Datasets will be released at \url{https://github.com/DonsetPG/multigrid-gnn}.
\end{abstract}

\section{Introduction} \label{sec:introduction}

\IEEEPARstart{F}{inite} element methods have been crucial in modeling, simulating and understanding complex systems. They have become essential tools for Computational Fluid Dynamics (CFD) \cite{HACHEM20108643} and are used in many fields, such as mechanics \cite{BOUCHARD20033887}, electromagnetics \cite{marioni:hal-01649660}, or fluid-structure interaction \cite{FSICFL}). CFD tools have greatly improved efficiency, safety, and performance in various systems, while also reducing costs and environmental impact. This has led to continuous research by both academics and industries to enhance algorithms and methods for more accurate and effective CFD simulations.

While meshes are the natural support for CFD, they have not been the first focus of the Machine Learning (ML) community. The success of Convolutional Neural Networks (CNN) in image processing \cite{imagenet, alexnet} prompted their direct application to CFD. One significant application involves mapping meshes or velocity and pressure fields into images to exploit such CNNs. \cite{Tompson2016} conducted 3D-fluid simulations employing a CNN that forecasts subsequent images based on preceding ones. Similarly, both \cite{Thuerey2018} and \cite{Chen2019} utilized a U-net architecture to predict pressure and velocity fields given solely a shape as input. \cite{Chu_2017} applied a Generative Adversarial Nets (GAN) to simulate 3D flows.

\begin{figure*}[!t]
  \centering
  \includegraphics[width=0.88\textwidth]{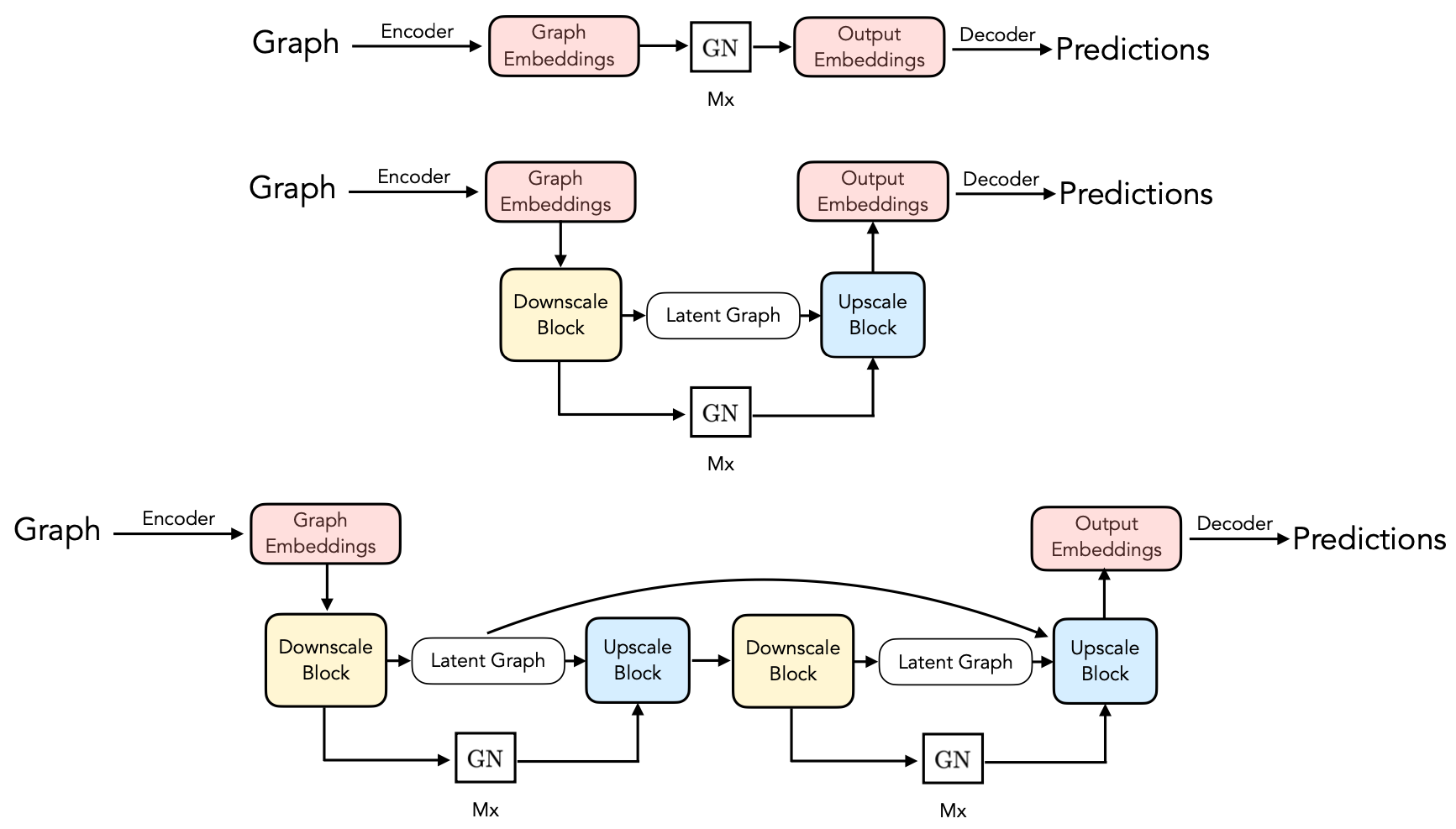}
\caption{
(top) The Encode Process Decode architecture (or MGN) from \cite{pfaff2021learning} with $M$ message passing steps. (middle) Our V-cycle model, with a depth of 1 and $M$ message passing steps in-between the DownScale and UpScaling blocks.
    (bottom) Our best-performing model, which consists of a W-cycle with Message Passing steps in-between.
} 
  \label{fig:full_model}
\end{figure*}

Still, the idea of using meshes directly as inputs for neural networks remains a natural approach, for which Graph Neural Network (GNN) \cite{gnn} can be leveraged. With the introduction of Message Passing GNN by \cite{Battaglia2018}, \cite{sanchezgonzalez2020learning} constructed a framework based on GNNs that made it possible to process unstructured grids or meshes directly. Based on this approach, \cite{pfaff2021learning} achieved state-of-the-art results on multiple CFD datasets, albeit restricted to small meshes (under 4000 nodes). To overcome this limitation, \cite{Yang2022amgnet, taghibakhshi2023mggnn,lino2021simulating} employed multiple graph coarsening stages, while \cite{Fortunato2022} built and operated with two graphs of different refinement stages from the start. Returning to CNNs, this MultiGrid approach can be put in parallel with U-net architectures \cite{ronneberger2015unet,lin2017feature}.

Beginning with Natural Language Processing (NLP), Transformers \cite{vaswani2023attention} achieved state-of-the-art results by replacing CNN and Recurrent layers with Self-Attention and Multi-Layer Perceptron (MLP). They now achieve state-of-the-art results in Computer Vision and Image Generation \cite{dosovitskiy2021image, arnab2021vivit} as well. Transformers have been applied to GNNs before \cite{yun2020graph,müller2023attending,veličković2018graph} to process the features of the graph nodes solely. \cite{shi2021masked} also introduced a Self-Attention mechanism to select the most important nodes of a graph.

Deep Learning architectures are now bigger and bigger and demand hundred of millions of labeled data. Unsupervised learning, particularly pre-training on unlabeled data, has emerged as a powerful technique for mitigating the costs associated with labeled data. The Cloze task, introduced by \cite{taylor1953cloze}, where missing words in sentences are inferred from the remaining context, has emerged as a cornerstone of this approach. \cite{Devlin2018} (BERT) pioneered the application of this framework to NLP, inferring masked tokens from the surrounding sentence. Similarly, \cite{he2021masked} introduced this approach for pre-training large networks processing images. While \cite{Hu2019} and \cite{Tan2022} attempted to adapt these methods for Graph Neural Networks, their efforts were limited to reconstructing node features or edges.

Driven by these analyses, we present a new model combined with a new  training method for CFD datasets. We also demonstrate that our results hold on meshes larger than on previous datasets by an order of magnitude (3k nodes to 30k nodes).

\begin{enumerate}
    \item Our model merges the approaches from \cite{Battaglia2018} and \cite{veličković2018graph}, using Self-Attention as the node-processing function in Message Passing blocks. This leads to a reduction of the all-rollout RMSE of 15\% on the \dataset{CylinderFlow} dataset from \cite{pfaff2021learning}.
    \item Our model goes further than both \cite{Fortunato2022} and \cite{lino2021simulating} by dynamically pruning our mesh based on Self-Attention, thus proposing a solid GNN-based multigrid approach. This leads as well to a reduction of the all-rollout RMSE of 15\%.
    \item We present a new self-supervised training method for GNN, based on BERT \cite{devlin2019bert} where a subset of nodes are removed from the initial graph. This change in training-paradigm itself leads to a reduction of the all-rollout RMSE of 25\% on every dataset.
\end{enumerate}

We conduct a comprehensive ablation study of model architecture, parameters, and regularization methods on the dataset introduced by \cite{pfaff2021learning}. Additionally, we train our models on a much more challenging dataset, both in terms of mesh size and dynamics complexity.

The present contribution introduces a model (see Figure \ref{fig:full_model}) that outperforms the state-of-the-art on \dataset{CylinderFlow} (71.4 $\rightarrow$ 29.4, $\downarrow$ 58\%) , \dataset{DeformingPlate} (16.9 $\rightarrow$ 4.5, $\downarrow$ 73\%) and \dataset{BezierShapes} (335 $\rightarrow$ 212, $\downarrow$ 37\%). Our self-supervised method alone leads to significant gain (71.4 $\rightarrow$ 46.5, $\downarrow$ 34\% on \dataset{CylinderFlow}), aligned with those witnessed in NLP \cite{radford2018improving} and images \cite{he2021masked}, and we hope that they will enable more research in that direction.

The paper is organized as follows: the theoretical frameworks behind Message Passing, Multigrid and Attention-layers are presented in section \ref{sec:tf-framework}. The regularization techniques such as node masking and noise, as well as the hyper-parameters and the datasets used are detailed in section \ref{sec:training}. Then, a full ablation study is performed, and the results of our models are shown in section \ref{sec:results}. Finally, perspectives on future works are given. The base code used in this paper is available at https.



\section{Theoretical framework} \label{sec:tf-framework}

We consider a mesh as an undirected graph $G = (V,E)$, where $V$ are the nodes and $E$ the edges. 
$V = \{\mathbf{v}_i\}_{i=1:N^v}$ is the set of nodes (of cardinality $N^v$), where each $\mathbf{v}_i$ represents the attributes of node $i$.
$E = \{\left(\mathbf{e}_k, r_k, s_k\right)\}_{k=1:N^e}$ is the set of edges (of cardinality $N^e$), where each $\mathbf{e}_k$ represents the attributes of edge $k$, $r_k$ is the index of the receiver node, and $s_k$ is the index of the sender node.

In the following, we refer to $G^{1h}$ as the graph associated to the mesh with the initial mesh size. In section \ref{subsec:up-down}, we introduce $G^{nh}$ as the same graph but with a mesh coarsened $n$ times by a ratio of $0.5$ (\textit{e.g.} $G^{2h}$ has half the amount of nodes as $G^{1h}$). The coarsening procedure is described in section \ref{subsec:up-down-block}. Each edge feature is made of the relative displacement vector in mesh space $\mathbf{u}_{ij} = \mathbf{u}_i - \mathbf{u}_j$ and its norm $\lVert\mathbf{u}_{ij}\lVert$. Each node features $\mathbf{v}_i$ (such as the pressure, velocity) also receives a one-hot vector indicating the node type (such as inflow or outflow for boundary conditions, obstacles to denote where shapes are inside the domain, etc) and global information (viscosity, gravity) creating $\mathbf{x}_i$\footnote{We find that adding historical data by repeating $\mathbf{v}_i$ for previous time-steps does not improve the long term rollout RMSE.}. 

For the case of 3D datasets, we follow the same approach as \cite{pfaff2021learning} and also add world-edges with a certain collision radius $r_D$ (\textit{i.e.} for each pair of non-neighbour nodes, if their world distance is smaller than $r_D$, we add a fake edge between them).

\subsection{Overall architecture}
\label{subsec:architecture}
The model is made of an encoder (see \ref{subsec:encoder}), a processor (see \ref{subsec:processor}) and a decoder (see \ref{subsec:decoder}). The processor comprises a stack of $M$ blocks, each block being either a GraphNet block from \cite{Battaglia2018}, a downscale block, or an upscale block (\ref{subsec:up-down-block}). These blocks aim to process spatial information between nodes, utilizing edge features. The inclusion of upscale and downscale blocks enables us to employ a multi-grid approach, dynamically pruning and refining our mesh (see figure \ref{fig:up-down-block}).

\subsection{Encoder}
\label{subsec:encoder}
We encode nodes and edges features with 2 simple Multi Layer Perceptron (MLP) into latent vectors of size $p$, following the same approach as in \cite{sanchezgonzalez2020learning}. 

\begin{equation}
\begin{alignedat}{3}
        &\mathbf{e}_k &&= \text{MLP}(\mathbf{u}_{ij}, \lVert\mathbf{u}_{ij}\lVert) &&\hspace{1em} \forall k \in E,
        \\
        &\mathbf{v}_r &&= \text{MLP}(\mathbf{x}_r) &&\hspace{1em} \forall r \in V,
\end{alignedat}
\end{equation}

where the MLP is made of 4 layers of hidden dimension of size $p$, ReLU activation and Layer Normalization.

\subsection{Multi-grid processor}
\label{subsec:processor}
\subsubsection{Graph Net blocks}
Our Graph Net blocks derive from \cite{Battaglia2018} and is made of a Message Passing layer that updates both the node and edge attributes given
the current node and edge attributes, as well as a set of learnable parameters. We first update the edges, then process an aggregation function before updating the nodes.

\begin{equation}
\begin{alignedat}{3}
        &\mathbf{e}_k' &&= f^e(\mathbf{e}_k,\mathbf{v}_{r_k},\mathbf{v}_{s_k}) &&\hspace{1em} \forall k \in E
        \label{eq:edge_model} \\
        &\bar{\mathbf{e}}_r' &&= \displaystyle\sum_{e \in E_r'} e &&\hspace{1em} \forall r \in V \\
        &\mathbf{\Tilde{v}}_r &&= [\mathbf{v}_r,\bar{\mathbf{e}}_r'] &&\hspace{1em} \forall r \in V \\
        &\mathbf{v}_r' &&= f^v(\mathbf{\Tilde{v}}_r) &&\hspace{1em} \forall r \in V
\end{alignedat}
\end{equation}

where $f^e$ is a simple MLP and  $f^v$ is the graph multi-head self-attention layer from \cite{veličković2018graph}. Usually, $f^v$ is also an MLP, but we find that using a Self-Attention layer allows to simulate another step of interaction betwenn nodes and features without adding many parameters, and without an extra message passing step. Each node feature $\mathbf{v}_r'$ is defined as:

\begin{align}
\mathbf{v}_r' = \sigma\left(\frac{1}{K}\sum_{k=1}^K \sum_{j\in\mathcal{N}_r}\alpha_{rj}^k{\bf W}^k\mathbf{\Tilde{v}}_j\right) \forall r \in V
\end{align}

where $\sigma$ is a softmax function, $K$ the number of attention heads, $\mathcal{N}_r$ the direct neighbours of $\mathbf{v}_r$, $\bf W^k$ a set of learnable parameters, and $\alpha_{rj}^k$ are attention parameters defined as:

\begin{align}
\alpha_{rj}^k = \frac{\exp\left(\text{MLP}([{\bf W}^k\mathbf{\Tilde{v}}_r,{\bf W}^k\mathbf{\Tilde{v}}_j])\right)}{\displaystyle\sum_{p\in\mathcal{N}_r} \exp\left(\text{MLP}([{\bf W}^k\mathbf{\Tilde{v}}_r,{\bf W}^k\mathbf{\Tilde{v}}_p])\right)}
\end{align}

\subsubsection{UpScale and DownScale blocks}
\label{subsec:up-down-block}
\begin{figure}[!t]
  \centering
  \includegraphics[width=2.5in]{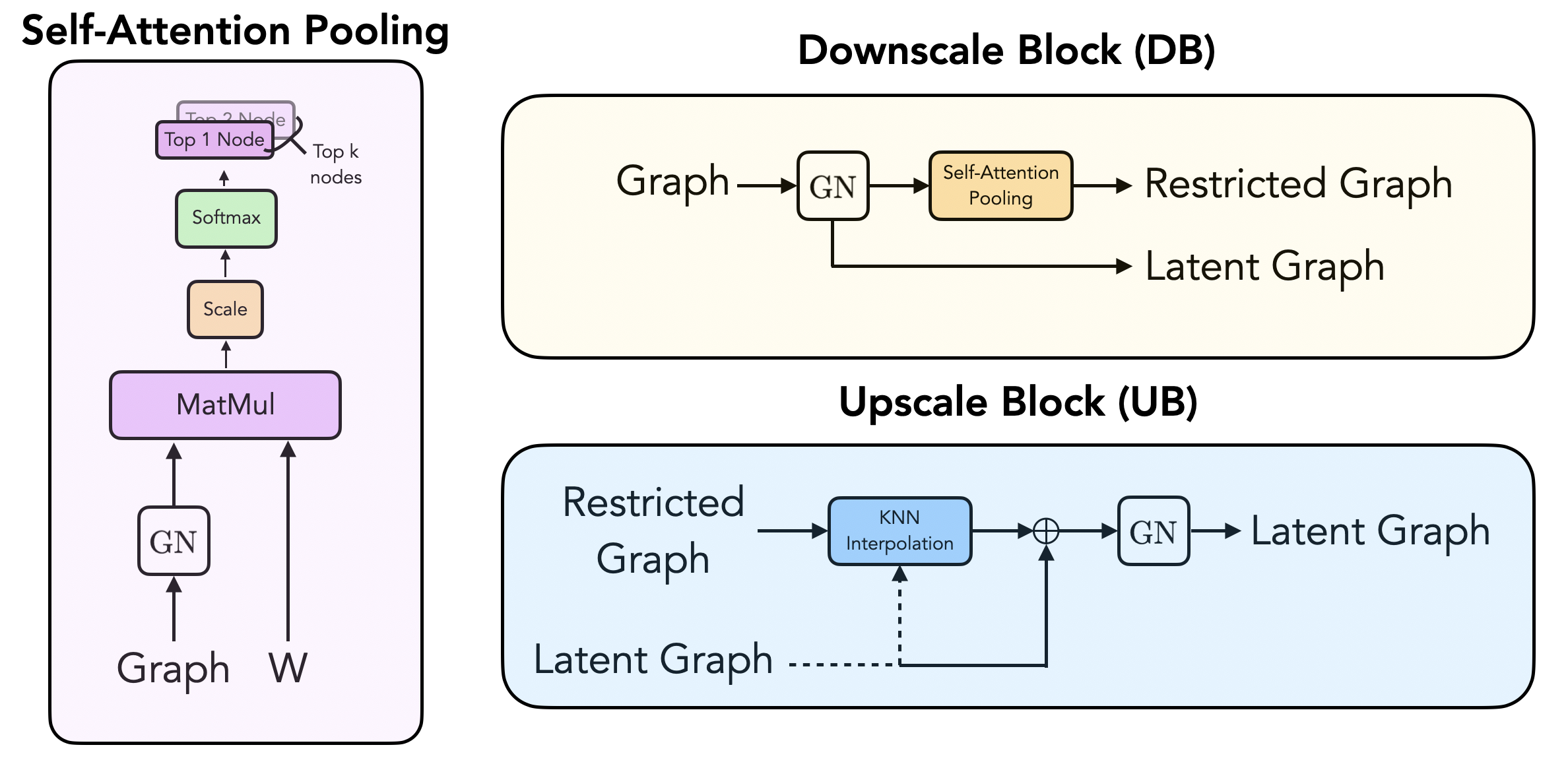}
\caption{
(left) Self-Attention Pooling block, to prune a graph by keeping the k nodes with the best Self-Attention Score. W is a set learnable parameters.
(top) Downscale block, keeping the top-k nodes based on their Self Attention Score. (bottom) Upscale Block based on a linear KNN interpolation.
} 
  \label{fig:up-down-block}
\end{figure}

We denote the pruning\footnote{We also tried the model with a re-meshing operation after the pruning, like a standard multigrid method. This gives similar results while increasing the inference time.} and refining operations as DownScale and UpScale blocks, respectively. Each block consists of a Message Passing block (before pruning or after refinement) and a scaling block. The blocks architectures are presented in figure \ref{fig:up-down-block}.

The DownScale block utilizes a Self-Attention pooling layer to rank each node and retain the top $k$ (in practice, we retain half the nodes). This layer follows the Scaled Dot-Product Attention architecture from \cite{vaswani2023attention}, applied to $\mathbf{x}$, the node features, as introduced in \cite{lee2019selfattention, knyazev2019understanding, shi2021masked}. We modify this layer by incorporating a Message Passing Block before computing the score:

\begin{align}
        \mathbf{y} &= \sigma \left( \frac{\mathbf{X}\mathbf{p}}{\| \mathbf{p} \|} \right)  \\
        \mathbf{i} &= \mathrm{top}_k(\mathbf{y})
\end{align}

where $\mathbf{X}$ are the nodes features after one step of Message Passing, $\mathbf{p}$ a set of learnable parameters, $\sigma$ a softmax function. In pratice, the DownScale block process a graph into a message passing step, computes $\mathbf{y}$, ranks each node according to $\mathbf{y}$ and then select the $k$ nodes with the highest score.

The UpScale block takes a fine and a pruned graph as inputs and interpolates the node features from the pruned graph onto the fine graph following the strategy proposed by \cite{qi2017pointnet}:

\begin{align}
y &= \frac{\displaystyle\sum_{i=1}^l w(x_i) x_i}{\displaystyle\sum_{i=1}^l
        w(x_i)} \textrm{, with } w(x_i) = \frac{1}{d(\mathbf{p}(y),
        \mathbf{p}(x_i))^2}
\end{align}

where $\mathbf{p}$ maps a node to its position, $d$ is a distance function, and $\{ x_1, \ldots, x_l \}$ the $l$-nearest points to $y$ (see figure \ref{fig:masking}).

The aforementionned blocks can be organised in cycles of various complexities: one DownScale block followed by an UpScale block (1D 1U) forms a V-cycle of depth 1. By adding more blocks, (2D 2U) forms a V-cycle of depth 2, and (1D 1U 1D 1U) forms a W-cycle of depth 1, as shown in figure \ref{fig:full_model}. 

\label{subsec:up-down}

\subsubsection{Why MultiGrid?}
We believe a technique for spreading information across multiple levels is needed. This is based on evidence about how information travel in graphs and insights from multigrid methods like \cite{Adams1999,Fortunato2022}. This emerges from two main considerations. Firstly, one step of message passing can't flow information for more than the length of a mesh edge. While we refine a mesh to enhance accuracy thus slows information spread. Second, as pointed out by \cite{luz2020learning} and \cite{Fortunato2022}, GNNs and Gauss-Seidel relaxations can both benefit from a multigrid approach as they only approximate errors locally. 

\subsection{Decoder}
\label{subsec:decoder}
To predict the node features at the following time step state from that at the current time step,  we add a decoding MLP to transform the latent $\mathbf{v}$ into the output features:

\begin{equation}
\begin{alignedat}{3}
        &\mathbf{y}_r &&= \text{MLP}(\mathbf{v}_r) &&\hspace{1em} \forall r \in V
\end{alignedat}
\end{equation}

where MLP follows the same architecture as in section \ref{subsec:encoder}.

\section{Training} \label{sec:training}

\subsection{Regularization}

\paragraph{Masked Training}
At each training step, we randomly sample 85\% of the nodes from the graph. It is important to note that we do not "remesh" the graph, \textit{i.e.} add extra edges to replace edges that were deleted because of masked nodes. This reduced graph is then passed into our model. The prediction is then upsampled onto the finer graph, following the same interpolation as the one used in the UpScale block. The goal of the model is still to predict the next step for the chosen set of features (but with 15\% less nodes). 

We then use the same model and the same dataset to continue with the training, but without node masking (and thus any interpolation).

\paragraph{Autoregressive Noise}
Since our model will make predictions autoregressively over long rollouts, its required to mitigate error accumulations.
Since during both pre-training and finetuning, the model is only presented with steps separated by at most one step $\Delta t$, it never sees accumulated noise
from previous predictions.
To simulate this, we use the same approach as \cite{sanchezgonzalez2020learning} and \cite{pfaff2021learning} and make our inputs noisy.
More specifically, we add random noise $\mathcal{N}(0,\sigma)$ to the dynamical variables.

We also experimented with Self-Conditioning, \textit{i.e.} after masked pretraining, during the finetuning phase, we compute the loss on $f(G_t)$ with a probability $p_{sc}$ and on $f(f(G_{t-1}))$ for the remaining steps. We find that while this method leads to improvements for diffusion models \cite{chen2023analog}, it does not improve the long term RMSE in our different datasets. 

\subsection{Parameters}

\paragraph{Network Architecture}All of the MLPs (the first Node and Edge encoder, the Decoder, and the Edge processor from our Graph Net blocks) are made of 2 hidden layers of size $128$ with ReLU activation functions. Outputs are normalized with a LayerNorm. The Node processor from our Graph Net block is composed of a single Attention layer from \cite{veličković2018graph} with 4 heads. DownScale blocks use a ratio of $0.5$ (\textit{i.e.} keeping half the nodes). 

In the case of MultiGrid, we precise the cycle type (V or W) as well as its depth. If not specified, all models are made of 15 Message-passing steps. \footnote{For the V-cycle, 4 of them take place before the DownScale block, 10 after, and one after the UpScale block (4D10U1 with U for UpScale and D for DownScale). For the W-cycle: 3D4U3D4U1.}

\renewcommand{\arraystretch}{1.1}%
\begin{table*}
\begin{center}
\begin{small}
\begin{tabular}[p]{|p{25.5mm}||c|c|c|c|c|c|c|}
	\hline
	\textbf{Dataset} &  
	\textbf{Solver} &
  \multiline{6.7mm}{\textbf{\centering\#\\nodes}}&
	\textbf{Dimension} &
	\multiline{6.7mm}{\textbf{\centering\#\\traj}}&
 	\multiline{6.7mm}{\textbf{\centering\#\\steps}}&
  \multiline{6mm}{\centering$\mathbf{\Delta t}$\\s}
  \\\hline\hline
    \dataset{CylinderFlow} & COMSOL & 2k & 2D Fixed Mesh & 100 & 600 & 0.01 \\\hline
    \dataset{DeformingPlate} & COMSOL & 1k & 3D Fixed Mesh & 100 & 400 & - \\\hline
    \dataset{Multiple Bezier} & Cimlib & 30k & 2D Fixed Mesh & 100 & 6000 & 0.1 \\\hline
\end{tabular}
\end{small}
\end{center}
\caption{Size and physical parameters of our different datasets. We also precise the origin of each dataset.}
\end{table*}
\renewcommand{\arraystretch}{1.0}%

\begin{figure*}[!t]
  \centering
  \includegraphics[width=0.88\textwidth]{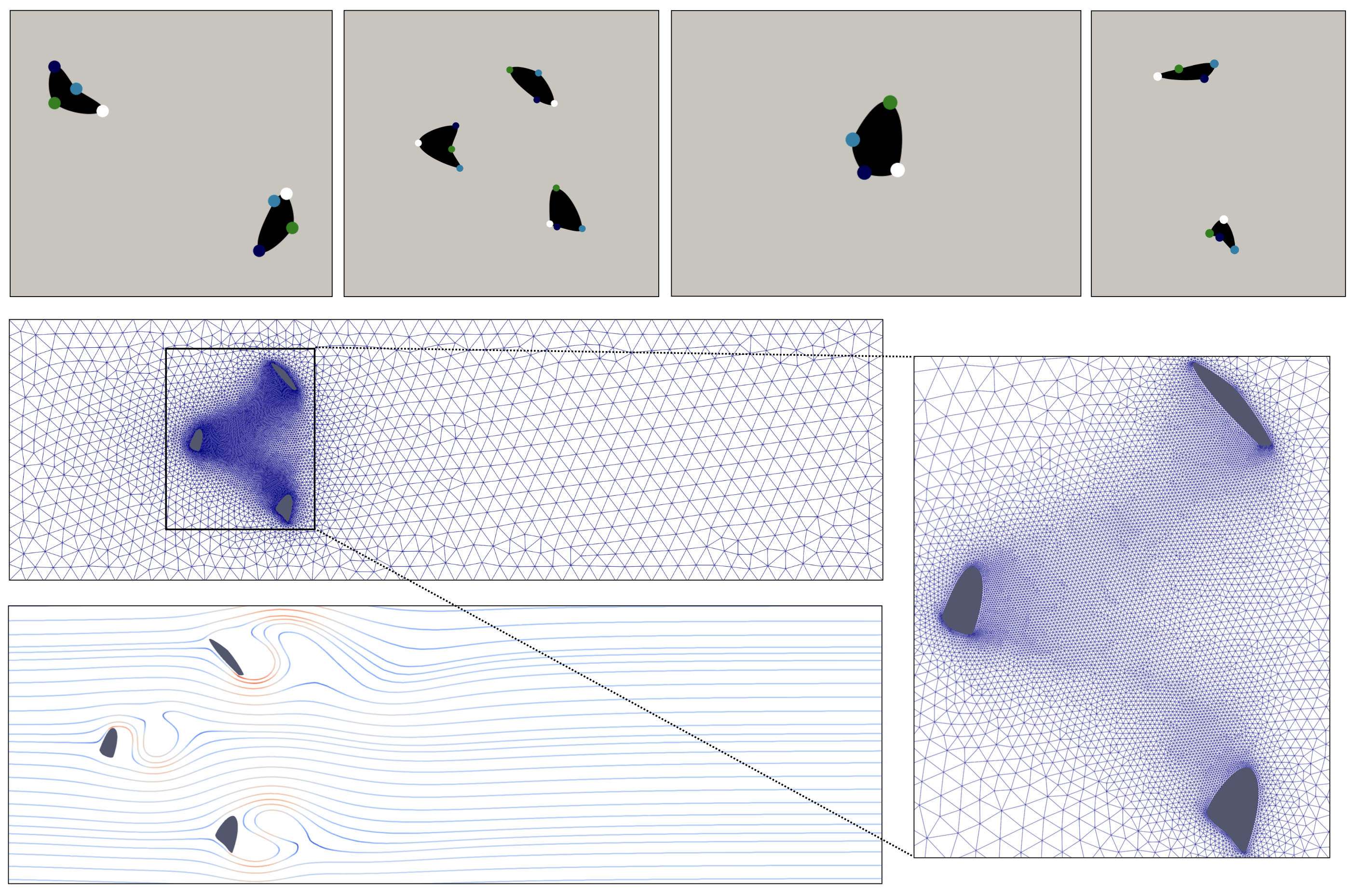}
\caption{
    \dataset{Bezier Shapes} dataset. (top) Sample of different shapes in the domain, with the control points used to build them. (middle and left) Example of a mesh. (bottom) Example of a velocity field.
} 
  \label{fig:bezier}
\end{figure*}

\paragraph{Training}We trained our models using an $L_2$ loss, with a batch size of $2$. We trained for 1M training steps, using an exponential learning rate decay from $10^{-4}$ to $10^{-6}$ over the last 500k steps.

For the masked training, we first train our model for 500k steps while masking 15\% of the nodes. We use an exponential learning rate decay from $10^{-4}$ to $10^{-6}$ over 250k steps. We then keep training the model for 500k more steps, while the full graphs. We start again with a learning rate of $10^{-4}$ before using the same schedule for the last 250k steps.

All models are trained using an Adam optimizer \cite{kingma2017adam}. 

\subsection{Datasets}

We conducted evaluations of the proposed model and its implementation across various applications, including structural mechanics and incompressible flows. Below, we provide an overview of the different use cases, the parameters utilized, and the simulation time step $\Delta t$. Each training set consists of 100 trajectories, while the testing set comprises 20 trajectories. The \dataset{CylinderFlow} and \dataset{DeformingPlate} datasets, sourced from the COMSOL solver, are originally described in \cite{pfaff2021learning}.

Our \dataset{Bezier Shapes} dataset simulates a incompressible flow around multiple random shapes (generated with a method from \cite{viquerat2020supervised}) at random positions (see figure \ref{fig:bezier}). The mesh contains the same physical quantities as the \dataset{Cylinder Flow} dataset with the same node types conventions (fluid nodes, wall nodes and inflow/outflow boundary nodes). The model also predicts changes in velocities and pressure per node. The Cimlib solver \cite{HACHEM20108643} was used to generate the trajectories. Meshes from this dataset are much larger than previous experiments, with on average 30k nodes.

\section{Results} \label{sec:results}

\begin{figure*}[t]
\vspace{-.2em}
\centering

\includegraphics[width=7.5in]{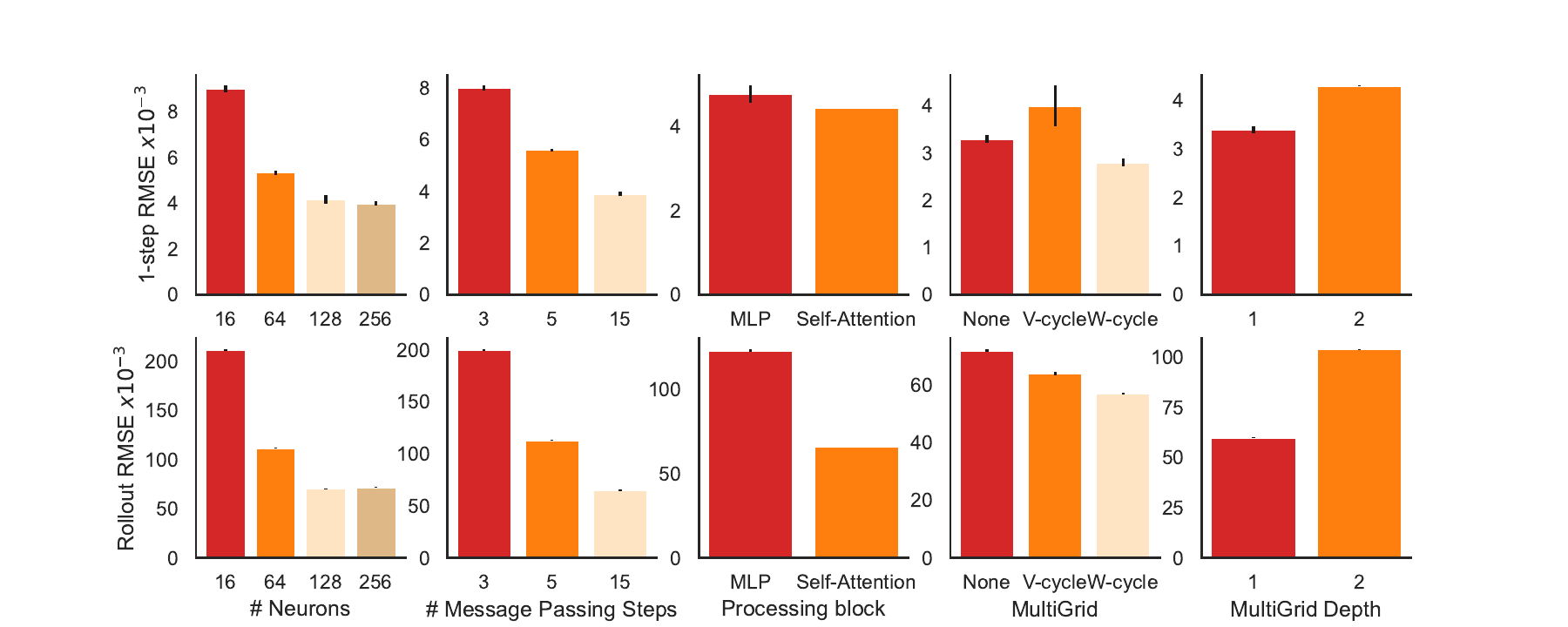}

\vspace{-.1em}
\caption{\textbf{Ablation study} on the Flow past a Cylinder Dataset. We tracked one-step RMSE and the RMSE averaged over the entire trajectory. All results are $\times10^{-3}$.
}
\label{tab:ablations} \vspace{-.5em}
\end{figure*}

We trained our best model on the 3 aforementionned datasets and compare it to 3 baseline models, including the state-of-the-art model from \cite{pfaff2021learning}. Our main finding is that each improvement proposed in this paper (node masking pre-training, attention layer, multigrid approach) offers substantial improvements, and that our best model outperform largely all existing baseline. It is also significantly faster than our in-house solver Cimlib (\cite{HACHEM20108643}). Notably, our Node masking approach could be generalized to much larger and complex dataset, for a fraction of the training cost. 

\subsection{Ablation Study}

\paragraph{Hyperparameters}
 We observed that increasing the number of neurons to 128 resulted in significantly improved outcomes. However, further increments beyond this threshold did not justify the associated increases in compute time and memory usage. Likewise, when considering the number of message passing steps, we found that exceeding 15 did not yield substantial improvements in comparison to the computation time.

We also observed that substituting $f^v$ with Self-Attention layers resulted in notable enhancements compared to a basic MLP, with a very small cost in terms of numbers of trainable parameters. Detailed results are presented in Figure \ref{tab:ablations}. However, it's noteworthy that on the \dataset{DeformingPlate} dataset, the Self-Attention layer contributed less to the improvements, with the majority of the enhancement stemming from the utilization of a Multigrid approach.

We  consistently observed these results across different model architectures, whether employing an Encode Process Decode architecture from \cite{pfaff2021learning} or a MultiGrid approach (utilizing both V or W-cycles).

In the following model, we adopt the parameters yielding the best results (mainly 15 Message Passing steps, with 128 neurons and with a Self-Attention layers in place of $f^v$). 

\paragraph{Multigrid}

We observed that transitioning from a simple Encode Process Decode model \cite{sanchezgonzalez2020learning, pfaff2021learning} to a MultiGrid model (either employing a V-cycle or a W-cycle) resulted in overall improvements. Additionally, we noted that W-cycle configurations consistently outperformed V-cycles across all datasets, aligning with the findings of \cite{Fortunato2022}.

\begin{figure}[t!]
  \centering
  \begin{subfigure}[b]{2.5in}
      \includegraphics[width=\textwidth]{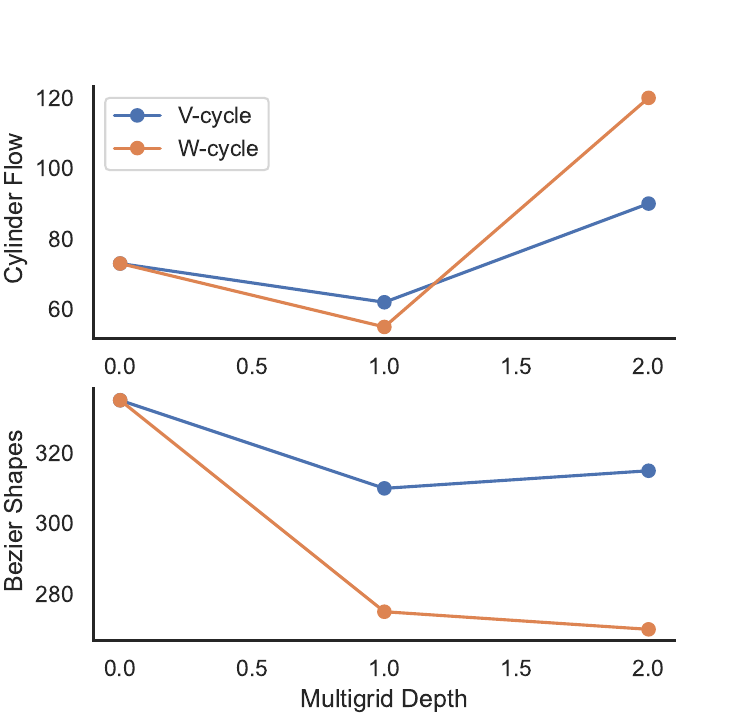}
      \label{fig:depth-results}
  \end{subfigure}
  \hspace{0.2em}
  \begin{subfigure}[b]{2.5in}
      \includegraphics[width=\textwidth]{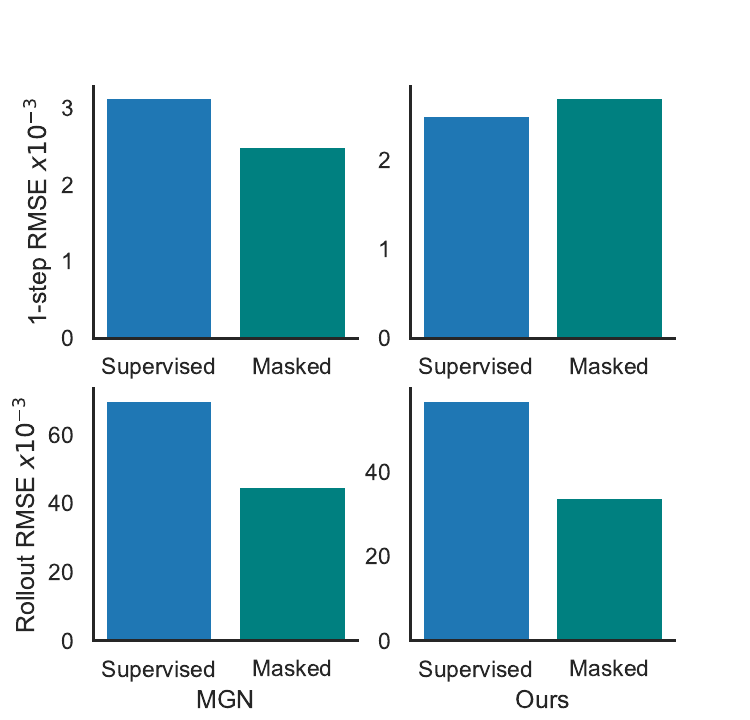}
      \label{fig:node-masking-results}
  \end{subfigure}
  \caption{(left) Comparison of the Rollout RMSE of our Multigrid approach for different depth. (right) Comparison of the same model trained with and without node masking. While no major difference can be found for 1-step RMSE, we see large gains for all-rollout.}
  \label{fig:other-ablation}
\end{figure}

However, we found that when the number of nodes was insufficient, moving from a depth-1 cycle to a depth-2 cycle did not always yield better results. For instance, with an average of 2k nodes, both V and W-cycles of depth 2 produced inferior results compared to their depth-1 counterparts (refer to Figure \ref{tab:ablations} and Figure \ref{fig:other-ablation}). On larger meshes (ranging between 20k and 30k nodes), we observed that deeper cycles yielded similar results.

\begin{table}[t!]
\begin{center}
\begin{tabular}[p]{lcc}

\toprule

            \textbf{Method} & \textbf{\dataset{Cylinder}-1} & \textbf{\dataset{Cylinder}-All}\\

\hline 

GCN \cite{kipf2017semisupervised} & 63.1 & 287 \\

U-Net \cite{Thuerey_2020} & 5.9 & 123 \\

MGN \cite{pfaff2021learning} & 3.3 & 71.4 \\

\hline

MGN + masking & \textbf{2.5} & 46.5 \\
MGN + attention & 3.3 & 58.1 \\

\hline 

V-cycle & 4.6 & 64 \\
W-cycle & 2.8 & 56.9 \\

\hline 

Ours (W-cycle + masking + attention) & 2.7 & \textbf{29.4} \\

\bottomrule
\hspace{5em}
\end{tabular}
\caption{\label{tab:main-res-cylinder}All numbers are $\times 10^{-3}$. \dataset{Dataset}-1 means one-step RMSE, and \dataset{Dataset}-All means all-rollout RMSE.}
\end{center}
\end{table}

\begin{table*}[t!]
\begin{center}
\begin{tabular}[p]{lcccc}

\toprule

            \textbf{Method} & \textbf{\dataset{Plate}-1} & \textbf{\dataset{Plate}-all} & \textbf{\dataset{Bezier}-1} & \textbf{\dataset{Bezier}-all} \\

\hline 

GCN \cite{kipf2017semisupervised} & 4.1 & 81.3 & - & - \\

MGN \cite{pfaff2021learning} & \textbf{0.07} & 16.9 & 27.7 &  335  \\

\hline

MGN + masking & 0.09 & 12.2 & 26.5 &  281  \\

\hline 

W-cycle & 0.17 & 8.1 & 24.1 &  275  \\

\hline 

Ours (W-cycle + masking + attention) & 0.11 & \textbf{4.5} & \textbf{17.9} & \textbf{212} \\

\bottomrule
\hspace{5em}
\end{tabular}
\caption{\label{tab:main-res}All numbers are $\times 10^{-3}$. \dataset{Dataset}-1 means one-stepp RMSE, and \dataset{Dataset}-All means all-rollout RMSE.}
\end{center}
\end{table*}

\subsection{Overall results}
Across three datasets with varying mesh sizes, physics dynamics, and complexity, our model consistently outperforms the current state of the art by a significant margin, ranging from 30\% to 50\% improvement (refer to Table \ref{tab:main-res-cylinder} and \ref{tab:main-res}). Notably, these improvements escalate to 50\%-75\% when utilizing node masking as a pre-training method (see Figure \ref{fig:other-ablation}). This underscores the fact that while node masking may not directly enhance 1-step RMSE performance, it encourages models to grasp the underlying physics intricacies instead of solely relying on extrapolation from a large visible set of nodes.

\begin{figure*}[!t]
  \centering
  \includegraphics[width=0.98\textwidth]{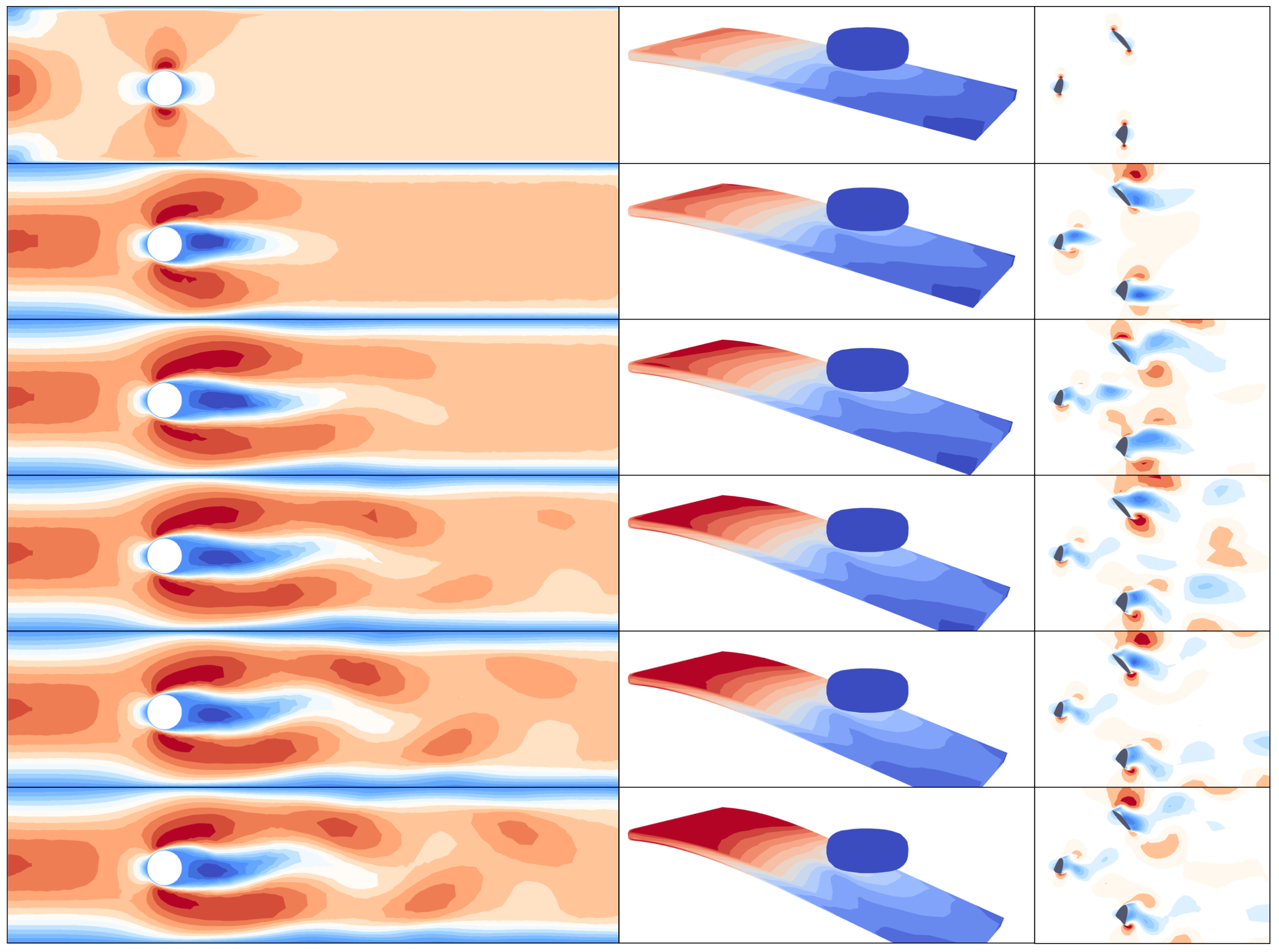}
\caption{
Prediction of our models on the 3 datasets. We display one frame every 25 time-steps. 
} 
  \label{fig:main-results-paraview}
\end{figure*}

We also find that using an Attention-based MultiGrid approach allows the model to process important information much quickly. The node selection closely follows the vortex created in \dataset{Cylinder}. In \dataset{DeformingPlate}, one layer follows the obstacle and the constraint on the plate, while the second selects the nodes moving the most within the plate (see Figure \ref{fig:node-selection}).

This approach also shows that combining different dynamic coarsening layers lets the model focus on different aspect of the graph, at different time of the spatial processing.

\begin{figure}[!t]
  \centering
  \includegraphics[width=2.5in]{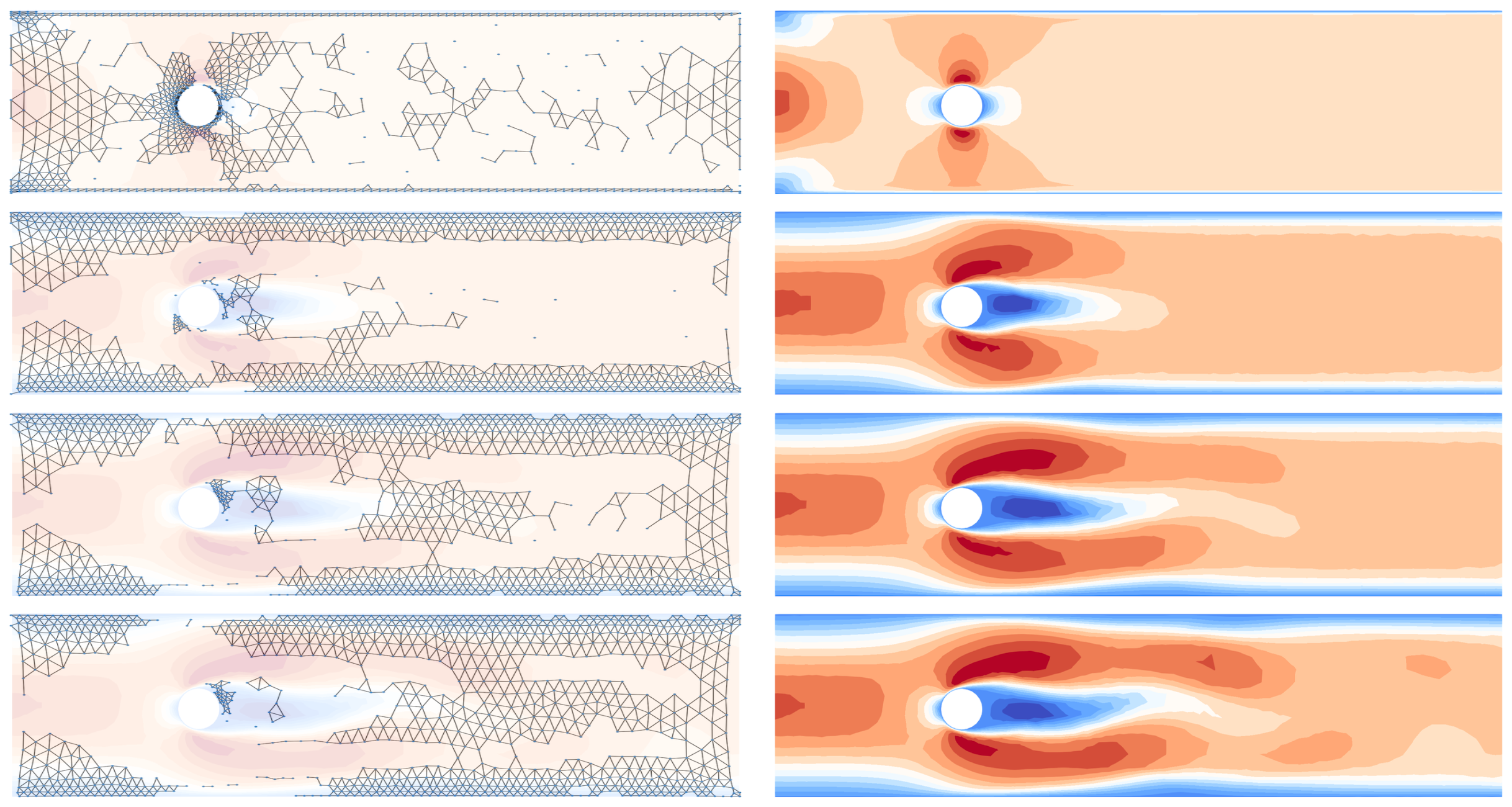}
  %
%
\caption{
Node selected by the Attention layer on the \dataset{CylinderFlow} dataset by a V-cycle multigrid model. We display one frame every 25 time-steps and keep the original mesh for the sake of vizualisation. 
} 
  \label{fig:node-selection}
\end{figure}

\begin{figure}[!t]
  \centering
  \includegraphics[width=2.5in]{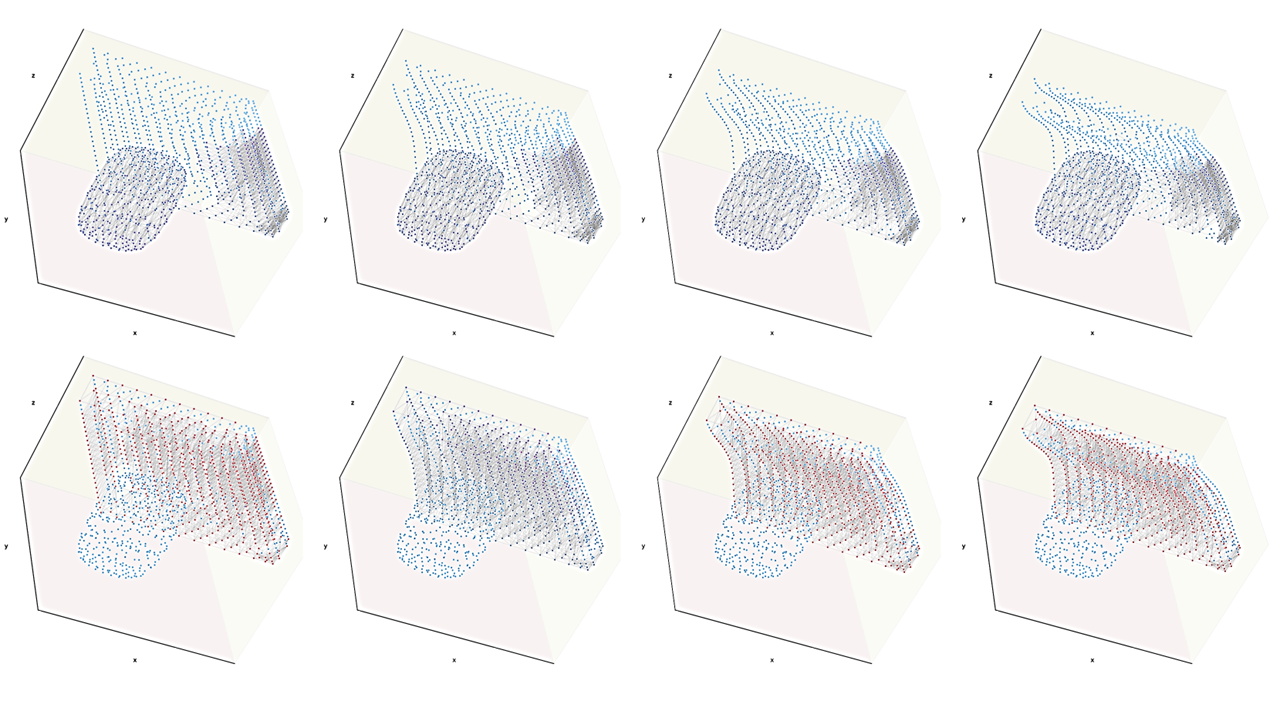}
  %
%
\caption{
Node selected by the Attention layer on the \dataset{DeformingPlate} dataset by a W-cycle multigrid model. We display one frame every 25 time-steps.
} 
  \label{fig:node-selection}
\end{figure}

\subsection{Generalization}

\begin{figure*}[!t]
  \centering
  \includegraphics[width=0.98\textwidth]{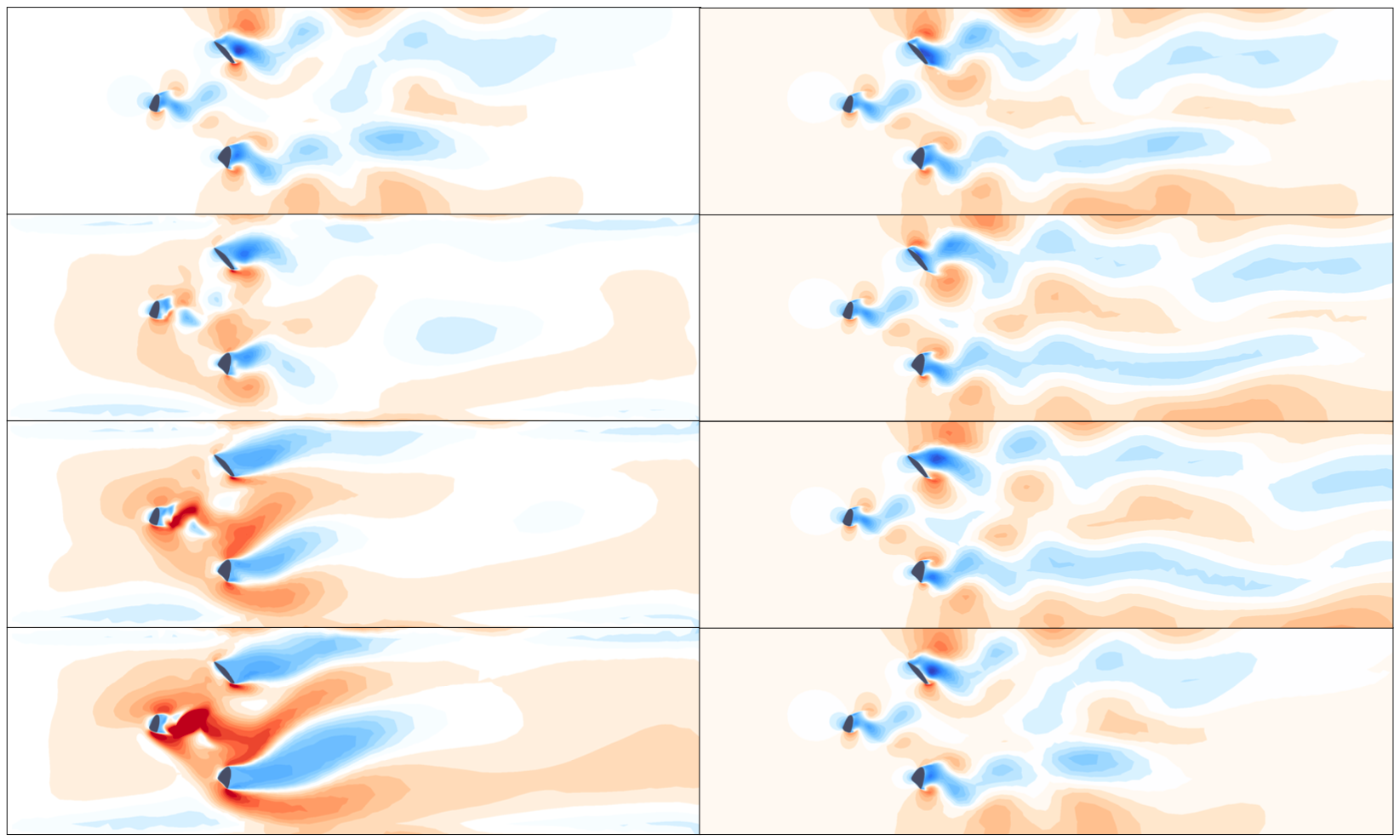}
  %
%
\caption{
Prediction from our best model trained on the \dataset{CylinderFlow} dataset. We showcase one prediction every 25 time-steps. (left) Prediction on a test case from the \dataset{BezierShape} dataset. (right) Ground-truth frames from \dataset{BezierShape}.
} 
  \label{fig:generalization-cylinder}
\end{figure*}

\begin{figure*}[!t]
  \centering
  \includegraphics[width=0.98\textwidth]{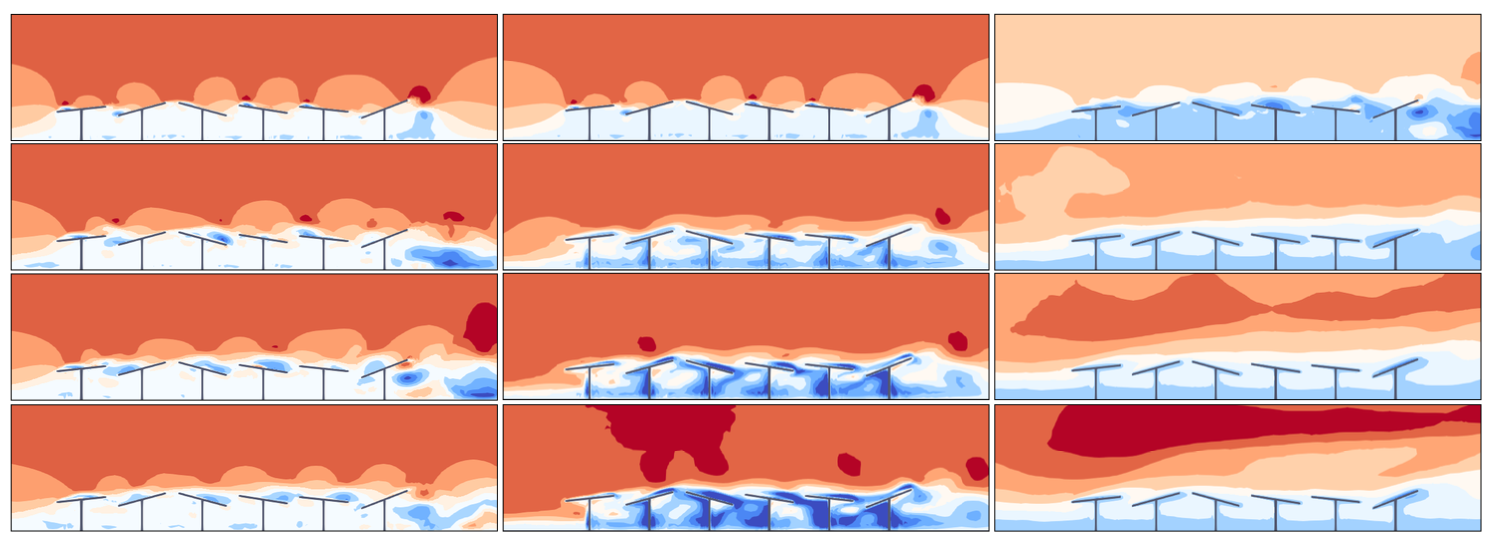}
  %
%
\caption{
Prediction from our best model trained on the \dataset{CylinderFlow} dataset. We showcase one prediction every 25 time-steps. (left) Ground-truth. (middle) Predictions from a model trained on \dataset{BezierShape}. (right) Predictions from a model trained on \dataset{CylinderFlow}.
} 
  \label{fig:generalization-cylinder-panels}
\end{figure*}

We noticed that our models exhibit strong generalization capabilities beyond the distribution of a specific dataset, maintaining performance consistency across similar domains, shapes, and meshes. This observation aligns with the findings reported in \cite{pfaff2021learning}. For samples really different from the training distribution, results can be coherent but much less accurate. For example, a model trained on \dataset{CylinderFlow} still produces good results on a test case from \dataset{BezierShapes}, with a close to ground-truth vortex for the middle shapes, and more averaged one for the shapes around it (see Figure \ref{fig:generalization-cylinder}). Similarly, a model trained on \dataset{BezierShapes} yields very good results on test cases from \dataset{CylinderFlow}.
On a much more difficult test case (in terms of mesh refinement, shapes and boundary conditions), our model struggles to get enough details or simply average a plausible flow over the domain (see figure \ref{fig:generalization-cylinder-panels})

While the results are not convincing at the moment, we believe this kind of generalization tasks are meaningful to understand if a model learns only the dataset distribution, or a form of physics. 
 
\section{Conclusion}

In conclusion, this study rigorously evaluated the performance of a novel model across three diverse datasets, benchmarking it against three baseline models, including the current state-of-the-art from \cite{pfaff2021learning}. Indeed, substantial improvements were offered by each enhancement introduced in this paper, namely, the node masking pre-training, attention layer incorporation, and multigrid approach. Notably, our best-performing model consistently outperforms all existing baselines by a significant margin. Moreover, it demonstrates remarkable efficiency, surpassing our in-house solver Cimlib in terms of speed.

Furthermore, our findings suggest that transitioning from a simple Encode Process Decode model to a MultiGrid model, particularly employing a W-cycle configuration, significantly enhances overall performance across datasets. While increasing the depth of cycles may not always lead to improved results, particularly with limited nodes, deeper cycles show promise on larger meshes.

Additionally, the proposed models exhibit strong generalization capabilities beyond dataset distributions. This suggests robustness and adaptability across various domains, shapes, and meshes, in line with the state-of-the-art methodologies.

In summary, our comprehensive evaluation, coupled with advancements in model architecture and training techniques, underscores the potential for significant strides in computational fluid dynamics and related fields. As we continue to refine and expand upon these methodologies, we anticipate further advancements in simulation accuracy, efficiency, and generalizability, paving the way for transformative applications in diverse scientific and engineering domains.

\paragraph{Acknowledgements}

Funded/Co-funded by the European Union (ERC, CURE, 101045042). Views and opinions expressed are however those of the author(s) only and do not necessarily reflect those of the European Union or the European Research Council. Neither the European Union nor the granting authority can be held responsible for them.
 
\bibliography{main} 
\bibliographystyle{IEEEtran}

\end{document}